\documentclass[lettersize,journal]{IEEEtran}
\usepackage{amsmath,amsfonts}
\usepackage{array}
\usepackage[caption=false,font=normalsize,labelfont=sf,textfont=sf]{subfig}
\usepackage{textcomp}
\usepackage{stfloats}
\usepackage{url}
\usepackage{verbatim}
\usepackage{graphicx}
\usepackage[center]{caption}
\usepackage{algorithm}
\usepackage{algpseudocode}
\def\BibTeX{{\rm B\kern-.05em{\sc i\kern-.025em b}\kern-.08em
    T\kern-.1667em\lower.7ex\hbox{E}\kern-.125emX}}
\usepackage{balance}

\begin{document}
\title{Improving Reliability of Fine-tuning with Block-wise Optimisation}

\author{Basel Barakat $^*$, and Qiang Huang $^*$ 
\thanks{Manuscript created January, 2023. $^*$ both authors contributed equally to this work. The authors from the the School of Computer science at the University of Sunderland, Sunderland, UK.}}

\markboth{Journal of \LaTeX\ Class Files,~Vol.XX, No.XX, September~2023}%
{Layer-wise fine-tune}

\maketitle

\begin{abstract}
Finetuning can be used to tackle domain-specific tasks by transferring knowledge learned from
pre-trained models. However, previous studies on finetuning
focused on adapting only the weights of a task-specific classifier
or re-optimising all layers of the pre-trained model using the new task data.
The first type of methods cannot mitigate the mismatch between a pre-trained model and the
new task data,
and the second type of methods easily cause over-fitting when processing tasks with limited
data.
To explore the effectiveness of fine-tuning, we propose a novel block-wise optimisation mechanism, 
which adapts the weights of a group of layers of a pre-trained model.
In our work, the layer selection can be done in four different
ways. The first is layer-wise adaptation, which aims to search for the most
salient single layer according to the classification performance. 
The second way is based on the first one, jointly adapting a small number of 
top-ranked layers instead of using an individual layer. 
The third is block based segmentation, where  
the layers of a deep network is segmented into blocks by non-weighting layers, such as
the MaxPooling layer and Activation layer.
The last one is to use a fixed-length sliding window to group layers
block by block. 
To identify which group of layers is the most suitable for finetuning,
the search starts from the target end and is conducted by freezing other layers
excluding the selected layers and the classification layers.  
The most salient group of layers is determined in terms of classification performance.
In our experiments, the proposed approaches are tested on an often-used dataset, $Tf\_flower$,
by finetuning five typical pre-trained models, VGG16, MobileNet-v1, MobileNet-v2, MobileNet-v3, and ResNet50v2, respectively. 
The obtained results show that the use of our proposed block-wise approaches can achieve better
performances than the two baseline methods and the layer-wise method.

\end{abstract}


\section{Introduction}\label{sec:intro}
\IEEEPARstart{T}he rapid development of deep learning technologies
has made it easy to construct and train complex neural networks
\cite{srivastava2015}. The deep structure of neural networks has thus
gained tremendous success. 
However one of their critical challenges is that
it needs large amounts of data. 
Training a model for a specific task on a limited data can lead to poor generalization due to over-fitting.
Although lots of data can be now collected online, data annotation is always an expensive and time-consuming task.
Therefore, more often in practice, one would fine-tune existing networks by continuing
training it on the new task data. This can
benefit many applications not having sufficient data
by transferring learned knowledge from multiple sources to a domain-specific task \cite{bansal2020, lih2020}.

Fine-tuning can process a pre-trained network in three different ways.
The first is to freeze all the weights of the pre-trained network,
but optimise only classifier layers using new task data.
The second is to optimise the weights of all layers. 
The last one is done by adapting the weights of a subset layers of the pre-trained model
as they would be more useful to learn dataset-specific features 
than other layers.

In our preliminary experiments, we noticed that randomly tuning few individual layers of a
pre-trained deep network can yield different performances, and occasionally outperform the first two fine-tuning methods mentioned
above. We thus hypothesised that fine-tuning
a group of layers may lead to some interesting results. 
Although fine-tuning a subset of layers seems to be more instinctively reasonable than the first two, 
it is not yet fully investigated how to determine which layers in the pre-trained network 
can have a greater contribution than other layers, when being tuned on new task data.
We therefore propose a novel framework using block-wise fine-tuning in this paper
and aim to explore an efficient way 
to find out the salient layers relevant to the features of new task data
and improve fine-tuning reliability.

The block-wise mechanism is conducted by dividing a deep network into blocks,
each of which consists of a group of layers.   
To find out the most salient block to the target of a new task, 
the block-wise fine-tuning can be implemented 
in four different ways in our work.
The first way is layer-wise fine-tuning, which aims to search for the most
salient individual layer according to the obtained accuracy on target data. 
The second way is based on the results of layer-wise adaptation. 
It aims to jointly fine-tune the weights of several
top-ranked layers instead of using only an individual layer since
multiple layers could benefit fine-tuning as more parameters are to be adapted.
The third way is block wise fine-tuning, where  
the layers of a deep network are segmented into blocks by non-weighting layers, such as
the MaxPooling layer and Activation layer.
The last one is to use a fixed-length sliding window to group layers
block by block.
The most salient block is identified in terms of the accuracy performance
obtained by fine-tuning a subset layers.
The details of our proposed approaches will be introduced in the next sections.

The rest of this paper is organised as follows: Section 2 introduces
the previous studies in relation to fine-tuning; the theoretical framework
is presented in detail in Section 3; Section 4 and 5 describe the data set
used in this work and experimental set-up; The results and analysis of our experiments are presented in Section 6, and finally conclusions are drawn in the last section.



\section{Related Work}\label{sec:related_work}

Within the framework of transfer learning and relying on
the architecture of a pre-trained model (PTM), fine-tuning can adapt
the model parameters on the target data
and has become one of the most promising deep learning techniques
in different research fields, such as computer vision (CV),
natural language processing (NLP), and speech processing.

\subsection{Fine-tuning in Computer vision}\label{subsec:ft_cv}
In computer vision community, the annual ImageNet Large Scale Visual Recognition Challenge (ILSVRC) \cite{imagenet}
provided multiple images sources and
has resulted in a number of innovations in the architecture,
such as VGG\cite{vgg16}, Inception\cite{inception, christian2016}, 
MobileNetV2\cite{mobilenetv2}, and ResNet50\cite{resnet50}.
By fine-tuning, these high-performing pre-trained models 
are now widely used in image generation\cite{yaxing2018, zhou2017},
image classification\cite{simon_cvpr2019, tanveer2021, jung2015, younmgin2021}, image caption\cite{jiamei2022, jaemin2022}, 
anomaly detection\cite{hamdi2020,rippel2021, wu2021}, image retrieval \cite{xu2018, radenovic2019}, etc.  
In these previous studies, the development of fine-tuning techniques in CV
can be found.
 
The first aspect of fine-tuning is layer wise adaptation.
Regarding the studies on the roles of hidden layers, Yosinski et al. \cite{yosinski2014} 
conducted empirical study to quantify the degree of generality and specificity
of each layer in deep networks. 
The related studies \cite{yosinski2014, zeiler2014} further claimed that the 
low-level layers extract general features and the high-level layers
extract task-specific features in a deep network. 
Since then, further works have been conducted to exploits the
role of each layer. In \cite{tajbakhsh2016}, Tajbakhsh et
al. showed that tuning only a few high-level layers is
more effective than tuning all layers. Guo et al. \cite{guo2019} 
proposed an auxiliary policy network that decides whether to
use the pre-trained weights or fine-tune them in layer-wise
manner for each instance. 
In \cite{younmgin2021}, Ro et al. proposed an algorithm that improves fine-tuning performance
and reduces network complexity through layer-wise pruning and auto-tuning of layer-wise learning rates. 
To further reinforce auto-tuning of layer-wise
learning rate, Tanvir et al. \cite{tanvir2022} proposed 
RL-Tune, a layer-wise fine-tuning framework for transfer
learning which leverages reinforcement learning (RL)
to adjust learning rates as a function of the target
data shift.

The second aspect of fine-tuning is in relation to
hyper-parameter optimisation.
This is because the increasing complexity of deep learning 
architectures slow training partly caused by ``vanishing
gradients''. In which the gradients used by back-propagation are
extremely large for weights connecting deep layers (layers near
the output layer), and extremely small for shallow layers (near
the input layer); this results in slow learning in the shallow layers\cite{bharat2015}.
So, Bharat et al. \cite{bharat2015} proposed a method to allow larger learning rates to compensate
for the small size of gradients in shallow layers.
Since then, various approaches have been explored for better regularization of
the transfer learning with effective hyper-parameter selection.
In \cite{kornblith2019} Kornblith et al. proposed a grid-search based
approach to search for better hyper-parameters, and Li et al. \cite{lih2020}
provided an elaborate guideline of learning rates and
other hyper-parameter selections.
Furthermore, Parker et al. 
proposed provably efficient online hyperparameter optimization 
with population-based bandits, which is found to be effective in optimizing RL training \cite{parker2020} .
To improve fine-tuning by using optimiser, Loshchilov et al. \cite{adamW} designed
two robust optimisers, SGDW and AdamW,
by combining SGD \cite{bottou2012} and Adam \cite{kingma2015} with decoupled weight decay.
The work in \cite{alexey2021, ananya2022} also explored
the use of two optimisers, SGD and AdamW, on ImageNet like domains in terms of fine-tuning accuracy.
Recently, \cite{kumar2022} found that large gaps in performance between SGD and AdamW
occur when the fine-tuning gradients in the first “embedding” layer are much
larger than in the rest of the model, and claimed that freezing only the embedding layer
can lead to SGD performing competitively with AdamW while using less memory.

The third aspect is neural architecture search (NAS),
aiming to adapt the architecture of a pre-trained network to match to
the characteristics of new task data.
Liu et al. \cite{liuc2018} used a sequential model-based optimization
to guide the search through the architecture of the network.
Pham et al. \cite{pham2018} proposed an efficient NAS (ENAS)
with parameter sharing, which focuses on reducing
the computational cost of NAS by reusing the trained
weights of candidate architectures in subsequent evaluations.
Lu et al. \cite{lu2021} proposed neural architecture transfer (NAT) 
to efficiently generate task-specific custom
NNs across multiple objectives.
Kim et al. \cite{kim2022} proposed to reduces the search cost using given architectural information
and cuts NAS costs by early stopping to terminate the search process in advance.
In \cite{tanveer2021}, Tanveer et al. developed differentiable neural architecture
search method by introducing a 
differentiable and continuous search space instead of a discrete
search space and achieves remarkable efficiency, incurring
a low search cost.

\subsection{Fine-tuning in Natural Language Processing}\label{subsec:ft_nlp}
Compared to CV, NLP models was typically more shallow and thus 
require different fine-tuning methods \cite{jeremy2018}.
In NLP, Mikolov et al. \cite{mikolov2013} proposed a simple transfer
technique by fine-tuning pre-trained word embeddings,
a model's first layer, but has had a large impact in practice and
is now used in most state-of-the-art models. To mitigate LMs' overfitting
to small datasets, Jeremy et al. \cite{jeremy2018} proposed
discriminative language model fine-tuning to retain
previous knowledge and avoid catastrophic forgetting.
In the last couple of years, large language models, such as GPT \cite{gpt2020} and BLOOM \cite{bloom2022}, 
were developed by using mask learning on large amounts of text data.
Given the size of these large language models, fine-tuning
all the model parameters can be compute and memory
intensive \cite{fedus2021}.
Some recent studies \cite{liang2021, hej2022} have
proposed new parameter efficient fine-tuning methods that
update only a subset of the model’s parameters.
As adversarial samples of new task are usually out-of-distribution,
adversarial fine-tuning fails to memorize all the robust and generic linguistic features already
learned during pre-training.
To mitigate the impacts caused by this, Dong \cite{dong2021} et al.
proposed to use mutual information to measure how well an objective model memorizes the useful features
captured before. Furthermore, Mireshghallah et al. \cite{Mireshghallah2022} 
empirically studied memorization of fine-tuning methods using
membership inference and extraction attacks as large models
have a high capacity for memorizing training samples
during pre-training.

\subsection{Fine-tuning in Speech Processing}\label{subsec:ft_sp}
For speech processing, fine-tuning can work not only for language model
adaptation \cite{huang2021, guilaume2022}, but also for tuning acoustic models 
\cite{violeta2022, tsiamas2022, vanek2019, peng2021}.
Fine-tuning language models in speech processing is
same as its use in NLP.
Guillaume et al. \cite{guilaume2022} developed a method using a transformer architecture
to tune a generic pre-trained representation model for phonemic recognition.
For acoustic model adaptation, Violeta et al. \cite{violeta2022} proposed an intermediate fine-tuning 
step that uses imperfect synthetic speech to close the domain shift gap between the
pre-training and target data. 
Tsiamas et al. \cite{tsiamas2022} proposed to use an efficient fine-tuning technique
that trains only specific layers of our system,
and explore the use of adapter modules for
the non-trainable layers. 
Peng et al. \cite{peng2021} used fine-tuning to 
learn robust acoustic representation to alleviate the
mismatch between a pre-trained model and new task data.
The similar work can be also found in \cite{haidar2021}, where
Haidar et al. employed Generative Adversarial Network (GAN) \cite{goodfellow2014}
to fine-tune a pre-trained model to match to the acoustic
characteristics of new task data.

\begin{figure*}
\centering
\includegraphics[scale=0.7]{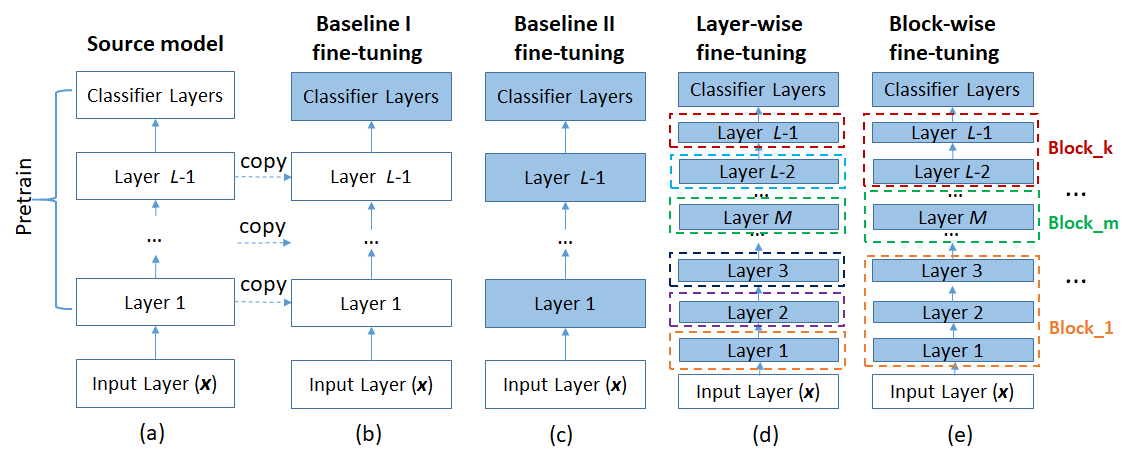}
\caption[center]{Architectures of source model and four target models: 
(a) source model 
(b) fine-tuning only classifier layers of source model , (c)fine-tuning all layers of source model,
(d)target model adapted by layer-wise fine-tuning, and (e) target model adapted by block-wise fine-tuning.}
\label{fig:th2}
\end{figure*}

\section{Theoretical Framework}\label{sec:theory}

\noindent Fine-tuning (\textbf{FT}) in this work is to adapt
the weights ($\textbf{W}$) of a group of layers ($\textbf{Ls}$) of 
a pre-trained deep neural network (\textbf{DN}) given
input data matrices $\textbf{X} = \{X_1,X_2,\ldots, X_M\}$ , where
$M$ is the number of training samples of a new task, 
and $X_m$ represents the $m$-th sample matrix.
The aim of the proposed approach is to find out the block $\textbf{B}_{Ls}$
most relevant to the target data. 
This can be represented by:
\begin{equation}
 \textbf{B}_{Ls,W} = \arg\max_{\textbf{Ls,W}} Accuray(\textbf{DN}(\textbf{X}))
\end{equation}
 
To attain the aim, we designed block-wise fine-tuning alongside the
work on layer-wise fine-tuning. Fig. \ref{fig:th2} shows the architectures
of source model and four target models for fine-tuning.
Fig. \ref{fig:th2}(a) is the pre-trained source model.
Fig.~\ref{fig:th2}(b) shows the target model (Baseline I) where only classifier
layers marked blue are to be adapted and the weights of other layers will be freezed. 
Fig.~\ref{fig:th2}(c)
is the target model (Baseline II), where all layers are to be re-optimised.
The two target models will be used as baseline models for a comparison in this paper.
Fig.~\ref{fig:th2}(c) and (d) show the architectures of
layer-wise and block-wise fine-tuning, respectively. The dash box means
when a layer or a group layers are being tuned, the weights of other layers, except classifier layers,
keep fixed.

\subsection{Layer-wise Fine-tuning}\label{subsec:layerwise}
As aforementioned in the first two sections, domain shift between
the pre-trained model and target data is the main reason why fine-tuning is needed.
The shift is actually accumulated by each individual layer of the pre-trained model.
So, our study in fine-tuning starts from layer-wise adaptation.
As shown in Fig.~\ref{fig:th2}(d), the weights of each layer will be adapted
to search for the layer, which can result in a salient improvement after its weights
are adapted on new task data.
The pseudo code in Algorithm \ref{alg:layerwise} shows 
the implementation of layer-wise fine-tuning, where
a while loop is used to search for the salient layer.
In our approach, the use of a small part of data, e.g. $10\%$, is to identify the most salient layer and thus
improve the fine-tuning efficiency.

\begin{algorithm}
\caption{Layer-wise fine-tuning (\textbf{FT})}\label{alg:layerwise}
\begin{algorithmic}[1]
\State    Load the weights, $\textbf{W}$, of a pre-trained model (\textbf{$\text{DN}$}) 
\State    Preprocess input data, $\textbf{X}$, and select 10\% data for training and 30\% data for evaluation 
\State    $N =$ \# layers of $\textbf{\text{DN}}$ 
\State    Initialise layer index, $i=1$, 
\State    Initialise accuracy array, $Acc=zeros(N)$
\State    \textbf{Training:}
\State    ~~While    {$i < N$}
\State    ~~~~~~tune the weights of the $i^{th}$ layer, $L_{i}$, and classifier
\State    ~~~~~~$Acc[i] = \textbf{FT}(\textbf{\text{DN}}(L_{i},C))$
\State    ~~~~~~$i = i + 1$
\State    ~~EndWhile
\State    ~~Identify the most salient layer, $L_{opt}$, using $Acc$
\State    \textbf{Evaluation:} 
\State    ~~Fine-tune the layer, $L_{opt}$, of the pre-trained model using all training data $(70\%)$ 
\State    ~~Evaluate the tuned model on the test data
  
\end{algorithmic}
\end{algorithm}

\subsection{Block-wise Fine-tuning}\label{subsec:blockwise}
In a deep neural network, several adjacent layers with a similar function can be
often grouped into a block. Tuning the weights of these layers in a block rather than all
layers is an efficient way to match the learned knowledge
to a specific task since only a relatively small part of 
parameters will be adapted. 
Fig. \ref{fig:th2}(d) shows the architecture of 
block-wise fine-tuning, starting from the input end of a deep network.

For block-wise fine-tuning, its critical step is to divide
the structure of a deep neural network into blocks. In our work,
the division can be done by non-weighting layers, such as
the Maxpooling layer and Activation layer.
In some typical deep neural networks, such as VGG16\cite{vgg16} and ResNet50\cite{resnet50},
there include not only Convolutional layers, but also 
MaxPooling, Batch Normalization, and Activation layers.
The convolutional layers in these models generally contain major parameters,
and only a relatively small number of parameters are from other layers.
This means Maxpooling layer, Batch Normalization layer, and even Activation layers
could be viewed as a delimiter to segment a deep neural network into blocks.

\begin{figure}[tbh]
\centering
\includegraphics[width=0.9\linewidth]{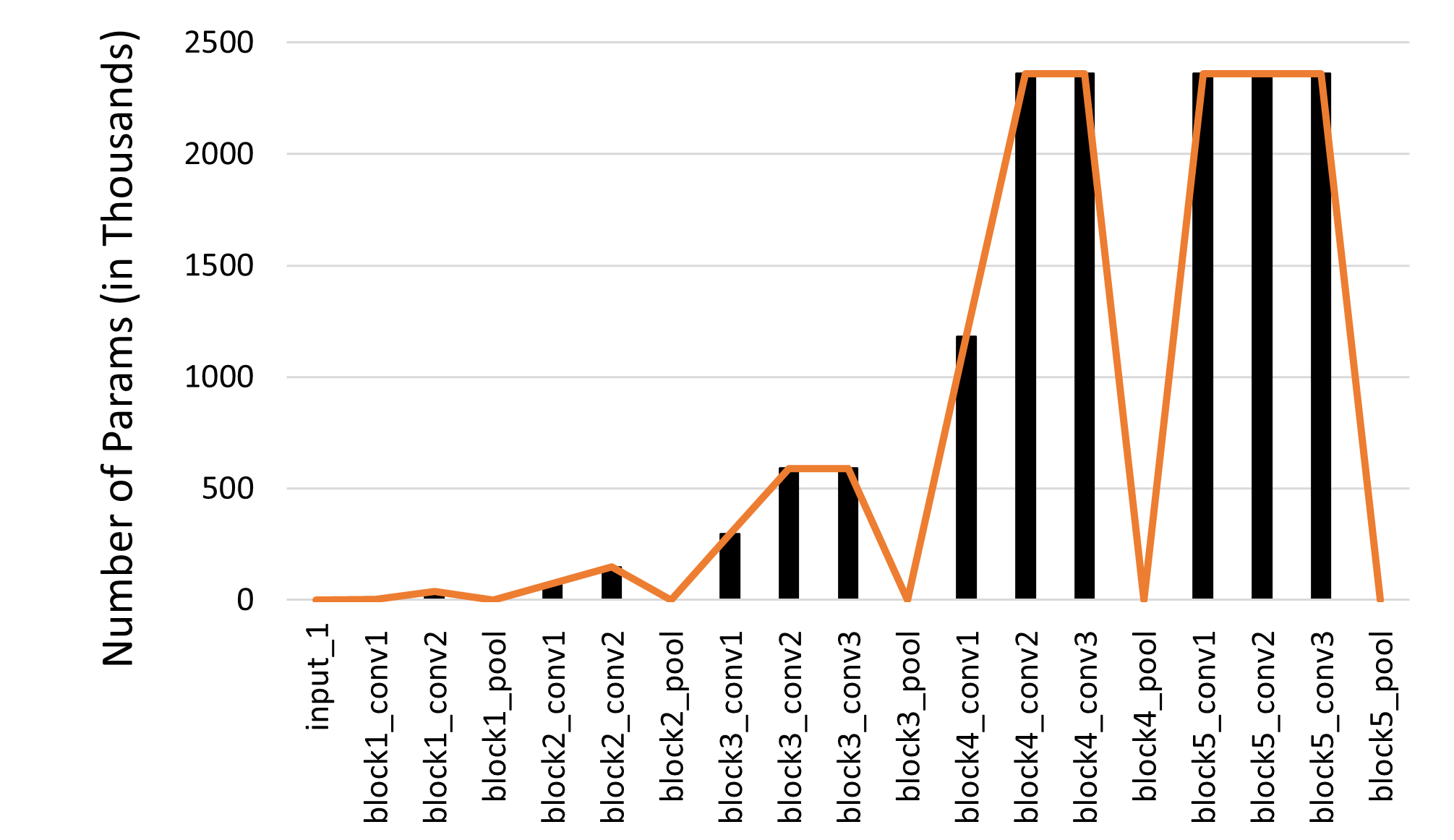}
\caption[center]{Distribution of the number of parameters of each layer of VGG16 model.}
\label{fig:vgg16_param}
\end{figure}

Fig. \ref{fig:vgg16_param} shows the number of parameters
of each layer of VGG16, a classical deep neural network for image classification.
In this figure, MaxPooling layers are corresponding to the valleys of the curve
since they are non-weighting layers.

%

\begin{algorithm}
\caption{Block-wise fine-tuning (\textbf{FT})}\label{alg:blockwise}
\begin{algorithmic}[1]
\State    Load the weights, $\textbf{W}$, of a pre-trained model ($\textbf{\text{DN}}$) 
\State    Preprocess input data, $\textbf{X}$, and select 10\% data for training and 30\% data for evaluation 
\State    $N =$ \# Blocks of $\textbf{\text{DN}}$ 
\State    Initialise layer index, $i=1$, 
\State    Initialise accuracy array, $Acc=zeros(N)$
\State    \textbf{Training:}
\State    ~~While  {$i < N$}
\State    ~~~~~~tune the weights of the layers in the $i^{th}$ block, $B_{i}$, 
\State    ~~~~~~and classifier
\State    ~~~~~~$Acc[i] = \textbf{FT}(\textbf{\text{DN}}(B_{i},C))$
\State    ~~~~~~$i = i + 1$
\State    ~~EndWhile
\State    ~~Identify the most salient block, $B_{opt}$, using $Acc$
\State    \textbf{Evaluation:} 
\State    ~~Fine-tune $B_{opt}$ of the pre-trained model using all training data (70\%)
\State    ~~Evaluate the tuned model on the test data
  
\end{algorithmic}
\end{algorithm}

Algorithm~\ref{alg:blockwise} shows the pseudo code of implementing
block-wise fine-tuning, where a while loop is run over blocks instead
of individual layers used in Algorithm~\ref{alg:layerwise}.
As an alternative, we can also use a sliding window instead of
non-weight layers to segment the long layer sequence into
block. The difference is that the length of the sliding window
is fixed and the number of layers in each block will thus be the same.
The window length is set to be three in our experiments.

\subsection{Block-wise Fine-tuning using ranked layers}\label{subsec:ft_ranked}
Unlike Algorithm \ref{alg:blockwise} to segment layers into blocks,
another block-wise fine-tuning approach is to select multiple layers,
whose contributions to target performance are top-ranked.
We thus make use of the evaluation performance on each individual
layer obtained by using Algorithm \ref{alg:layerwise}. In our experiments,
we select top three layers to make up a block.

\begin{algorithm}
\caption{Block-wise fine-tuning using ranked layers}\label{alg:blockwise_ranked}
\begin{algorithmic}[1]
\State    Load the weights, $\textbf{W}$, of a pre-trained model ($\textbf{\text{DN}}$) 
\State    Preprocess input data, $\textbf{X}$, to match to the input layer of $\text{DN}$ 
\State    $N =$ \# layers of $\textbf{\text{DN}}$ 
\State    Initialise layer index, $ii=0$
\State    ~~Initialise accuracy array, $Acc=zeros(N)$
\State    \textbf{Training:}
\State    ~~\textbf{ Implement Algorithm} \ref{alg:layerwise}
\State    ~~~Identify three top-ranked layers in terms of $Acc$  
\State    \textbf{Evaluation:}
\State    ~~fine-tune the selected three layers of pre-trained model using all training data $(70\%)$  
\State    ~~test the tuned model on the test data
\end{algorithmic}
\end{algorithm}

\section{Data set}

The algorithms were tested on the Tf\_flowers dataset \cite{tfflowers}.
The Tf\_flowers dataset is a collection of images of flowers, divided into five different classes: daisy, dandelion, roses, sunflowers, and tulips. There are a total of 3670 images in the dataset. The images are high quality color images resized to a resolution of $224\times224$ pixels, then we scaled RGB to 0-1 range. The dataset is intended for use in image classification tasks, and can be used to train machine learning models to recognize different types of flowers \cite{datasetapply}.

\section{Experimental set-up}

The training process for this experiment involves using the pre-trained models as a starting point, and then fine-tuning the models on the TensorFlow Flowers dataset. We used TensorFlow Keras library to build and train the models. The training process will involve using a small number of training samples, i.e., 10\% of the dataset, to identify the best performing layers (for Layer-wise FT), block (for Block-wise FT ), or Ranked block (for Block-wise with ranked layers FT). Then using the remaining training dataset (70\% of the whole dataset) to  evaluate the accuracy.

We have evaluated the performance of the algorithms on five commonly used pre-trained models, i.e., VGG16 \cite{vgg16}, MobileNet \cite{mobilenetv1}, MobileNetV2 \cite{mobilenetv2}, MobileNetV3 \cite{mobilenetv3}, and  ResNet50 \cite{resnet50}. The models were pre-trained on the imagenet wights.

 The classifier layers, we started with a Flatting layer \cite{kerasflat}, to which flattens the dimensions of the output to prepare it for use in Fully Connected (FC) layers \cite{Chollet}. Then a dense layers \cite{kerasdense}, with 128 units, followed by a dropout layer \cite{kerasdropout} with a rate of 0.5, to reduce the probability of overfitting. \cite{Srivastava}. Then two dense layers with  64 and 5 units respectively. The 64 units layer is activated by a `relu' function, and the final dense layer has five units, activated by `softmax' for the multi-class classification \cite{Goodfellow}.

The model is then compiled with loss function ``categorical cross-entropy", an optimizer ``Adam" with a learning rate of $ 5 \times 10^{-5}$ and metrics as ``accuracy". 
A ``Reduce LR On Plateau" callback is then defined, which reduces the learning rate when the validation accuracy plateaus. \cite{Chollet}.
The model is then trained on the training data with batch size of 4, 50 epochs and validation data and callbacks.

\section{Results and Analysis}
\begin{figure}[tbh]
\centering
\includegraphics[width=0.9\linewidth]{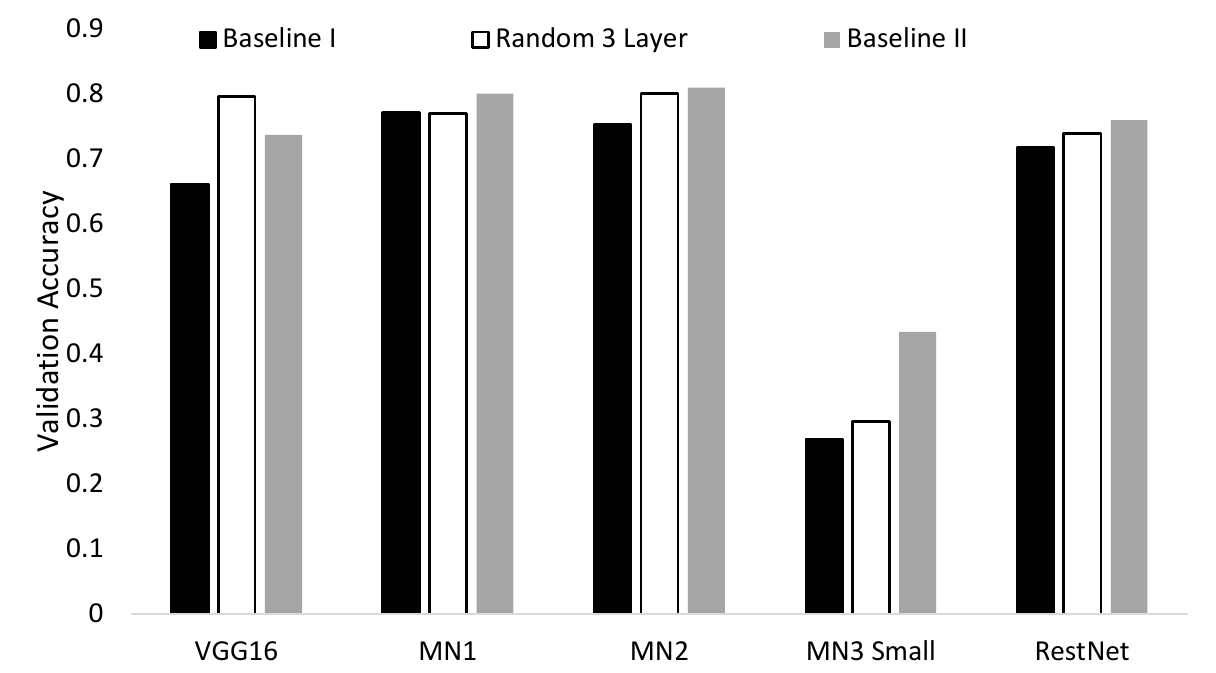}
\caption[center]{Comparison between the baseline models and FT three randomly selected layers using 10\% of the dataset.}
\label{fig:random}
\end{figure}

In this paper, we began our investigation by evaluating the performance of a Baseline I model, which only tunes the classifier layers, a Baseline II model, which tunes all layers, and a model that randomly tunes three layers using only 10\% of the dataset for training. As shown in Fig. \ref{fig:random}, the model with randomly chosen layers outperforms the Baseline I method in all tested models, and achieves a comparable performance with Baseline II.

\begin{figure*}
\centering
\includegraphics[scale=0.9]{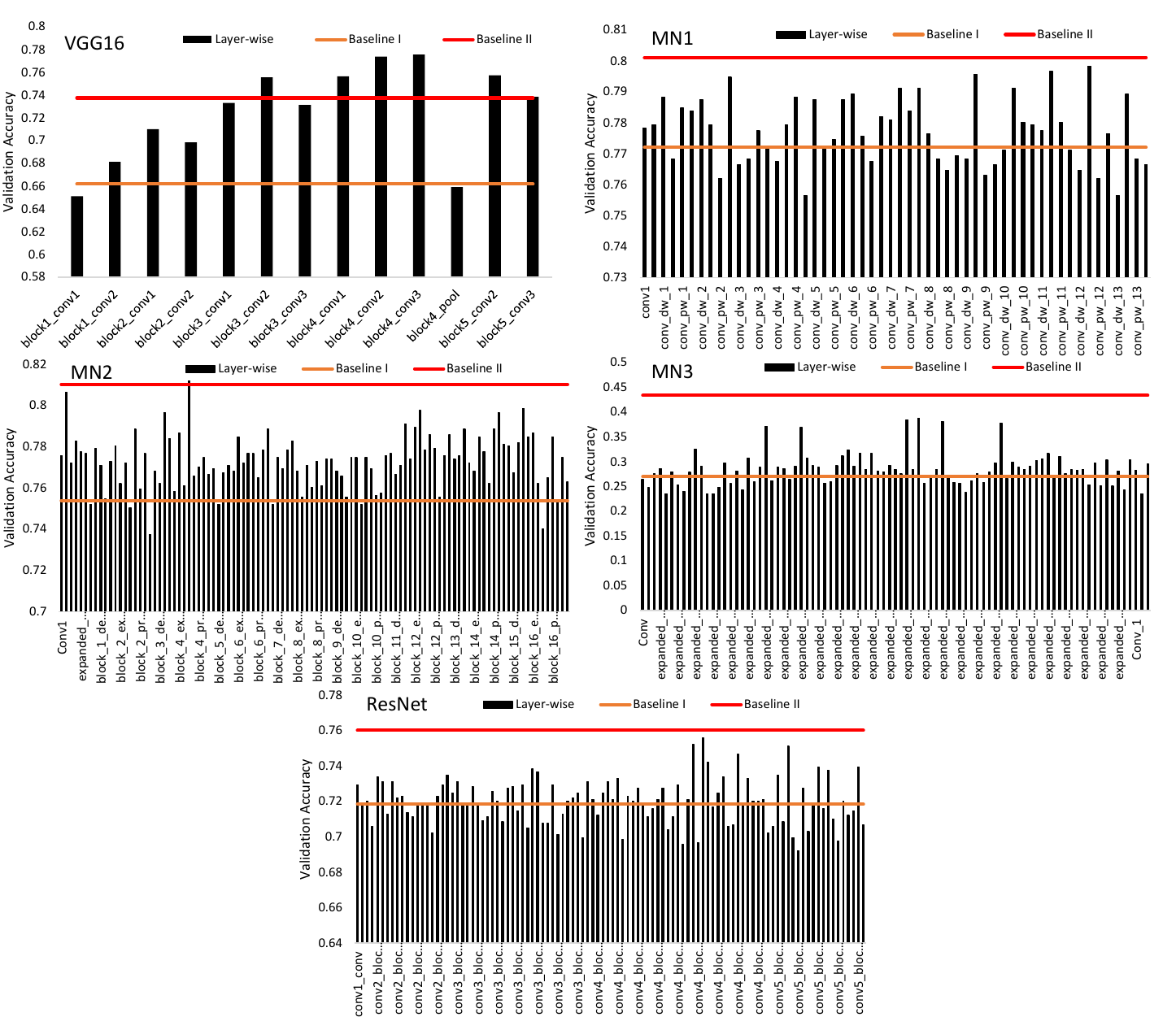}
\caption[center]{Layer-wise fine tuning accuracy using 10\% of the dataset for all tested models.}
\label{fig:D10SL}
\end{figure*}

Afterwards we investigated the performance of the layer-wise FT approach. As shown in Fig. \ref{fig:D10SL}, the classifier accuracy varies significantly between different layers. A very few layer are able to identify the classes with a high accuracy, even exciting both Baseline approaches, while the vast majority would not achieve high accuracy. 

\begin{figure*}
\centering
\includegraphics[width=0.9\linewidth]{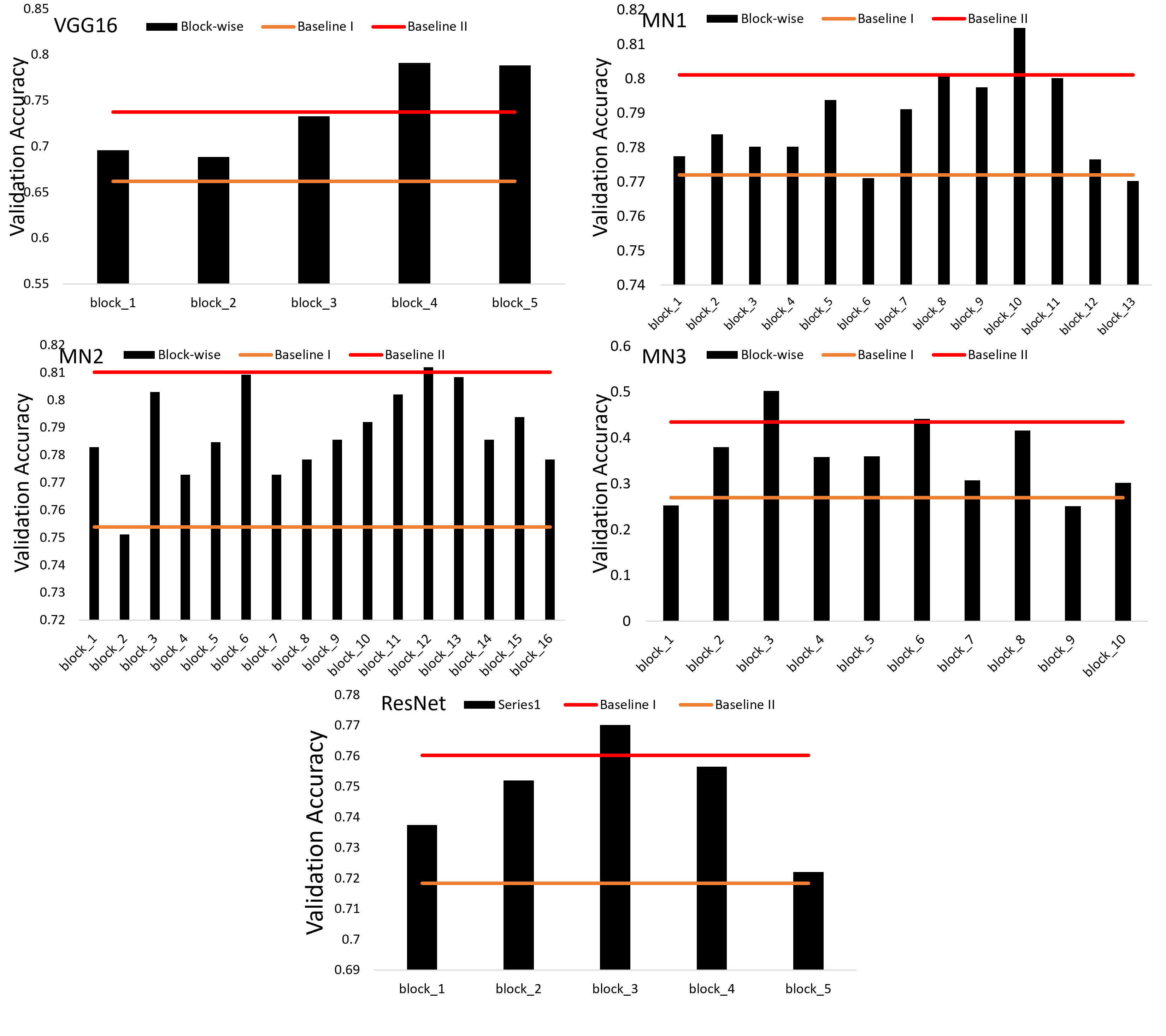}
\caption[center]{Block wise fine tuning accuracy using 10\% of the dataset. Showing the best performing block with a white filling with vertical dash.}
\label{fig:D10_Block}
\end{figure*}

For the block-wise FT, we implemented it using the pre-trained models with the 10\% of the dataset. The results shown in Fig. \ref{fig:D10_Block}. We can observe that the classification accuracy has been significantly improved with the use of block wise FT, and the accuracy has a certain observable pattern. We can also note that the highest accuracy blocks are not usually convective.   

\begin{figure*}[h]
\centering
\includegraphics[scale=0.9]{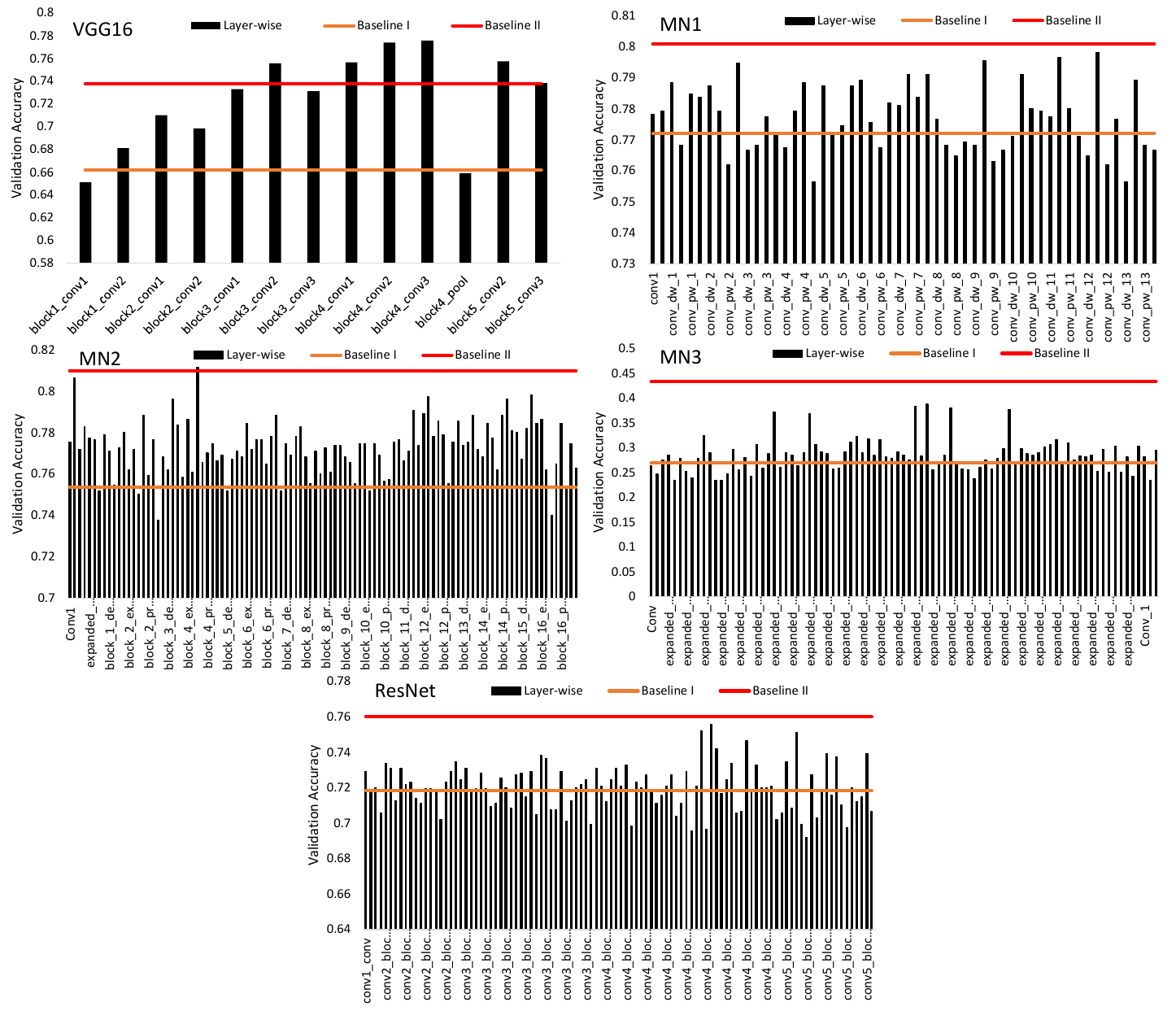}
\caption[center]{Sliding window fine tuning accuracy using 10\% of the dataset.}
\label{fig:D10_SW}
\end{figure*}

The next part of the investigation we evaluated the SW approach. Unlike the block-wise approach the SW approach all the blocks has a fixed number of trainable layers. Thus to asses the accuracy we had to loop through all the layers as shown in Fig. \ref{fig:D10_SW}. Although the number of trained layers in each `window' are relatively small, the achieved accuracy was comparable with the block-wise and the Baseline II approaches.  

\begin{figure}[tbh]
\centering
\includegraphics[width=0.9\linewidth]{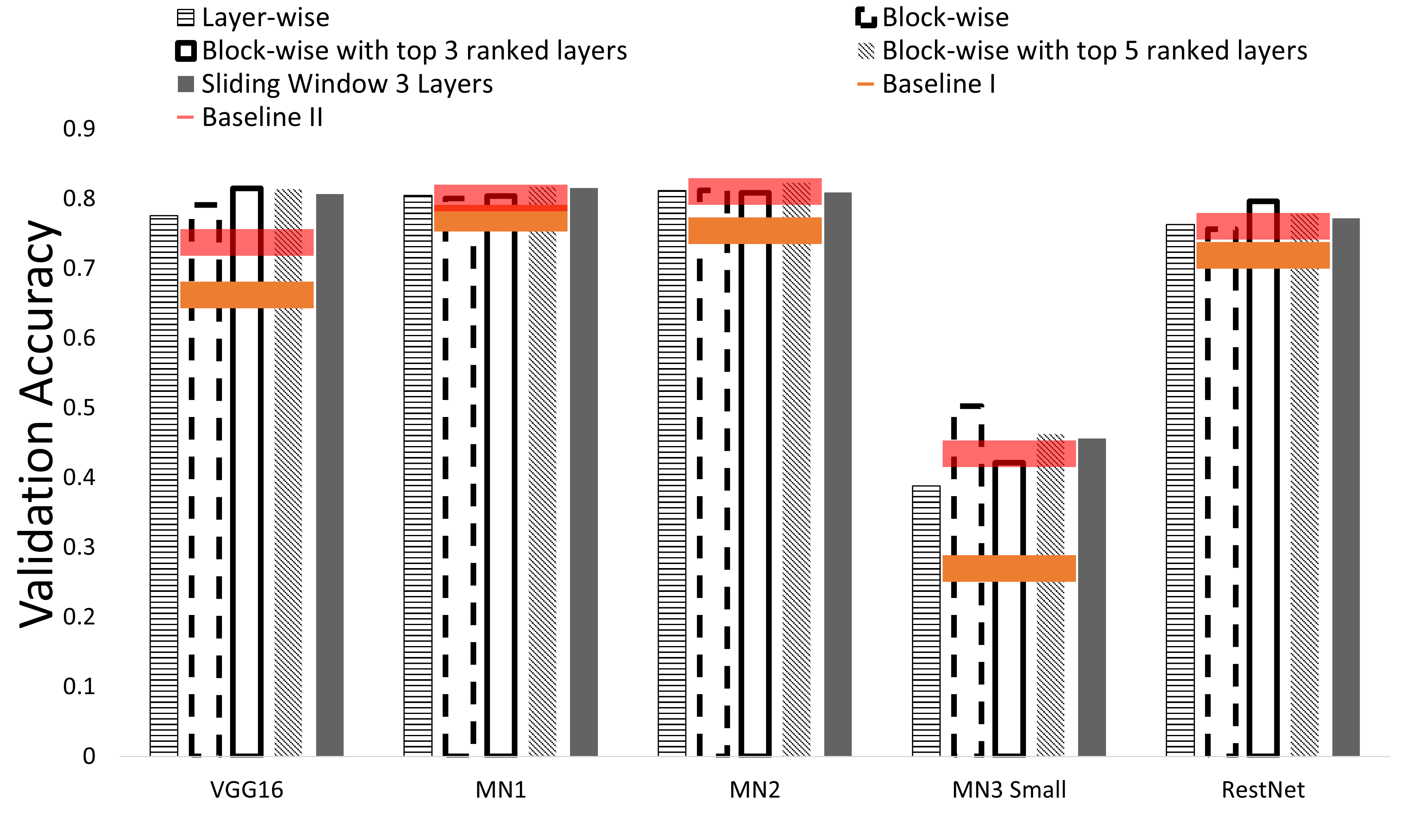}
\caption[center]{Fine Tuning validation accuracy using 10\% of the dataset. Showing the different approaches and the baseline performance.}
\label{fig:D10_summary}
\end{figure}

To summarise we have obtained the optimal layer, block, and `window' using only 10\% of the data set. As shown in Fig. \ref{fig:D10_summary}, all the tested approaches outperform Baseline I approach and few outperform Baseline II approach. Consequently we leveraged these findings to FT the models on the 70\% of the dataset.   

\begin{figure}[tbh]
\centering
\includegraphics[width=0.9\linewidth]{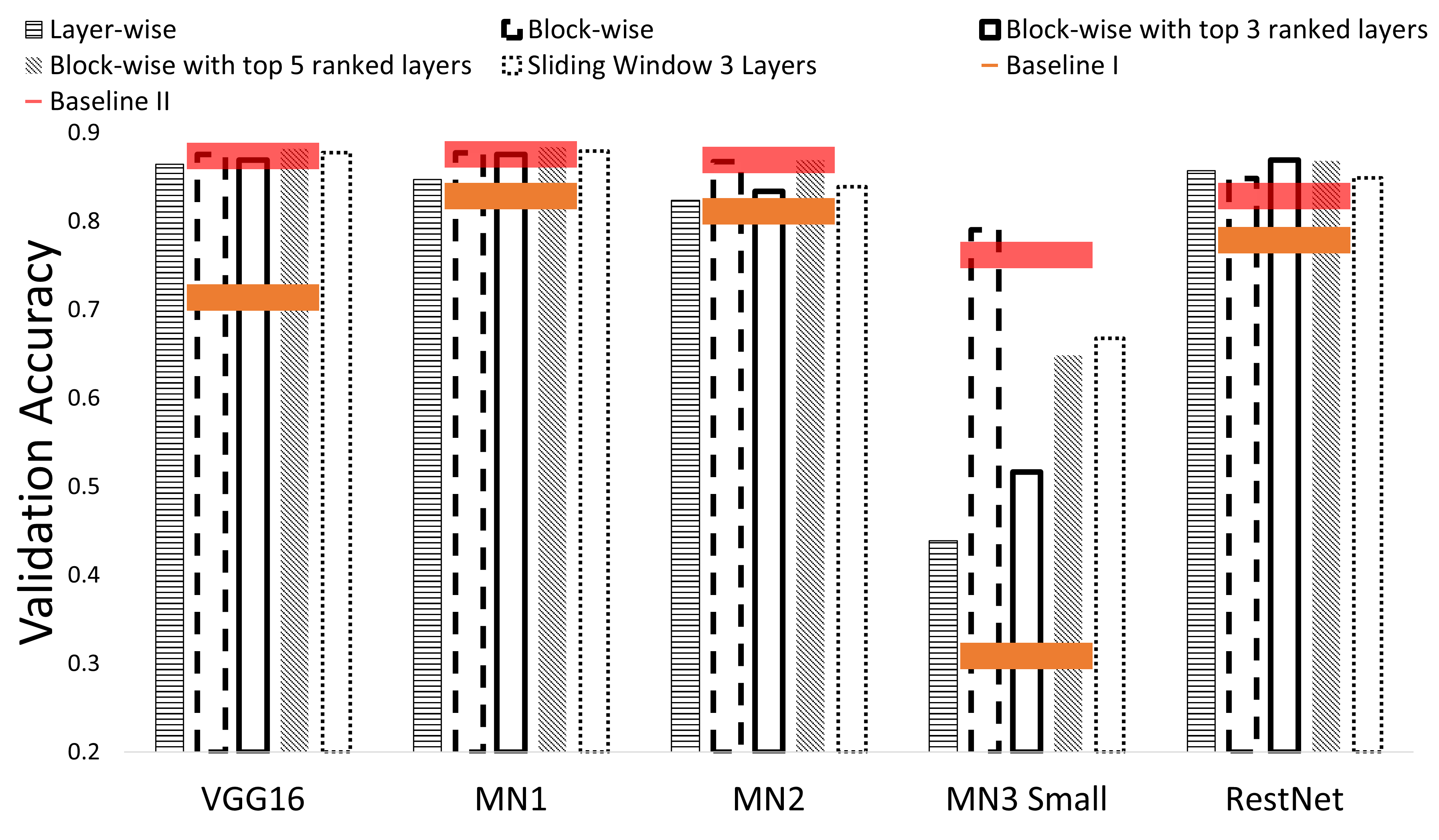}
\caption[center]{Block wise fine tuning accuracy using 70\% of the dataset. Showing the best performing block with a white filling with vertical dash.}
\label{fig:D70_summary}
\end{figure}

The fine-tinning accuracy using all the approaches using 70\% of the dataset is shown in Fig. \ref{fig:D70_summary}. We can observe that the block-wise approach had been consistent in achieving high accuracy. As it has always either score higher or equal to the Baseline II approach. To validate this findings we have repeated the experiments few time and results for the block-wise and the SW approaches are shown in Appendix\ref{sec:repeated}.

To evaluate the reliability and accuracy of the FT we evaluated the classification accuracy on all the models using all the baseline and proposed approaches, as show in Table \ref{tab:mean}. We can observe that the Block-wise approach has the highest average validation accuracy outperforming the Baseline II approach. Moreover it has a lower accuracy variance, thus yield a consistent reliable performance.  

\begin{table*}[]
\caption{Fine tuning validation accuracy showing both the mean and the variance of the tested approaches using 70\% of the data. LW refers to layer-wise, BW refers to block-wise, BWT3 refers to block-wise with the 3 top ranked layers, and BWSW refers to the sliding-window approach.}
\centering
\begin{tabular}{llllllll}\hline
          & \multicolumn{1}{c}{Baseline I} & \multicolumn{1}{c}{Baseline II} & \multicolumn{1}{c}{LW} & \multicolumn{1}{c}{BW} & \multicolumn{1}{c}{BWT3} & \multicolumn{1}{c}{BWT5} & \multicolumn{1}{c}{BWSW} \\ \hline
VGG16     & 0.713896     & 0.873751        & 0.864668           & 0.875568               & 0.86921                & 0.881926                 & 0.877384 \\
MN1       & 0.828338     & 0.875568        & 0.847411           & 0.877389               & 0.875568               & 0.883742                 & 0.879201 \\
MN2       & 0.811081     & 0.86921         & 0.823797           & 0.867393              & 0.833787               & 0.86921                  & 0.839237 \\
MN3 Small & 0.30881      & 0.762035        & 0.438692           & 0.790191              & 0.516803               & 0.648501                 & 0.667575 \\
RestNet   & 0.778383     & 0.828338       & 0.857402            & 0.848320               & 0.86921                & 0.868302                 & 0.849228  \\ \hline
Variance  & 0.046867     & 0.002364       & 0.033797            & \textbf{0.001318}      & 0.024095               & 0.010383                 & 0.007806  \\
Mean      & 0.688102     & 0.84178        & 0.766394            & \textbf{0.851771}      & 0.792916               & 0.830336                 & 0.822525 \\ \hline              
\end{tabular}\label{tab:mean}
\end{table*}

\section{Conclusions and future work}

Our preliminary experiments suggest that randomly tuning individual layers of a pre-trained deep network can yield intriguing results, with few even outperforming traditional fine-tuning methods. We proposed a novel framework using block-wise fine-tuning to explore an efficient way to find the salient layers relevant to the features of new task data and improve fine-tuning reliability. Our proposed approach includes four different methods for determining the most salient block: layer-wise fine-tuning, layer-wise adaptation, block-wise fine-tuning, and a fixed-length sliding window. This work can be extended in several approaches such as building an automated FT approach, that can automatically identify the best performing blocks and FT them.

\begin{appendices}\label{sec:repeated}
\section{Repeated experiments}
\begin{figure}
\centering
\includegraphics[width=0.9\linewidth]{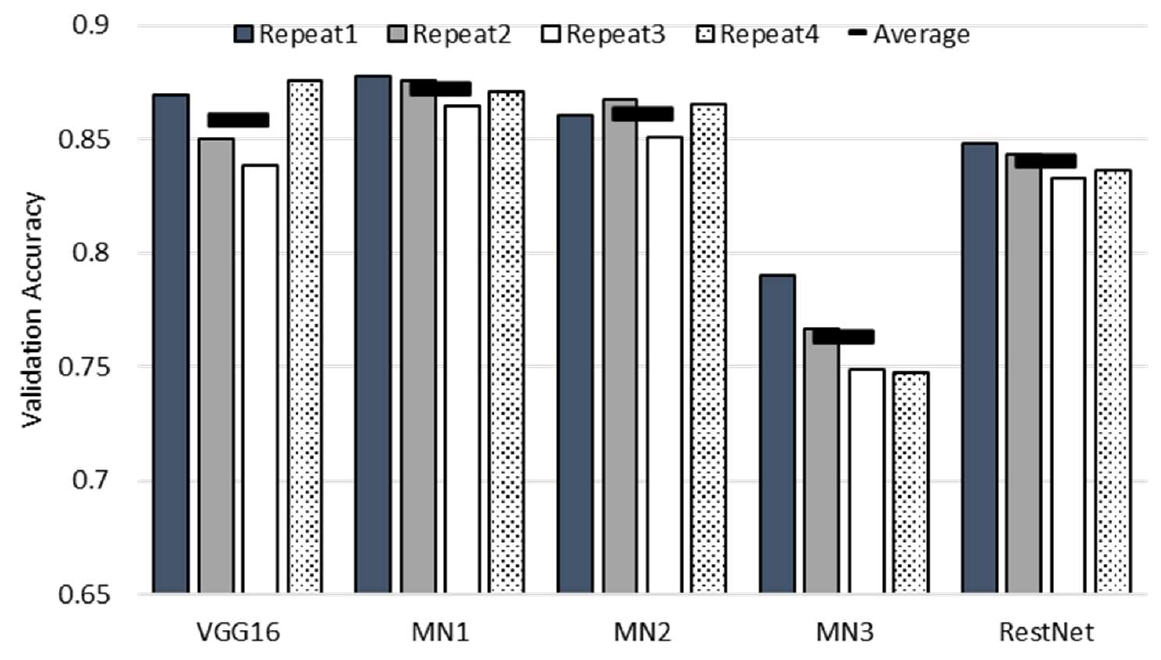}
\caption[center]{Block wise fine tuning accuracy using 70\% of the dataset.Showing the four repeated experiment and the resulting average.}
\label{fig:D70_Block_rep}
\end{figure}

\begin{figure}
\centering
\includegraphics[width=0.9\linewidth]{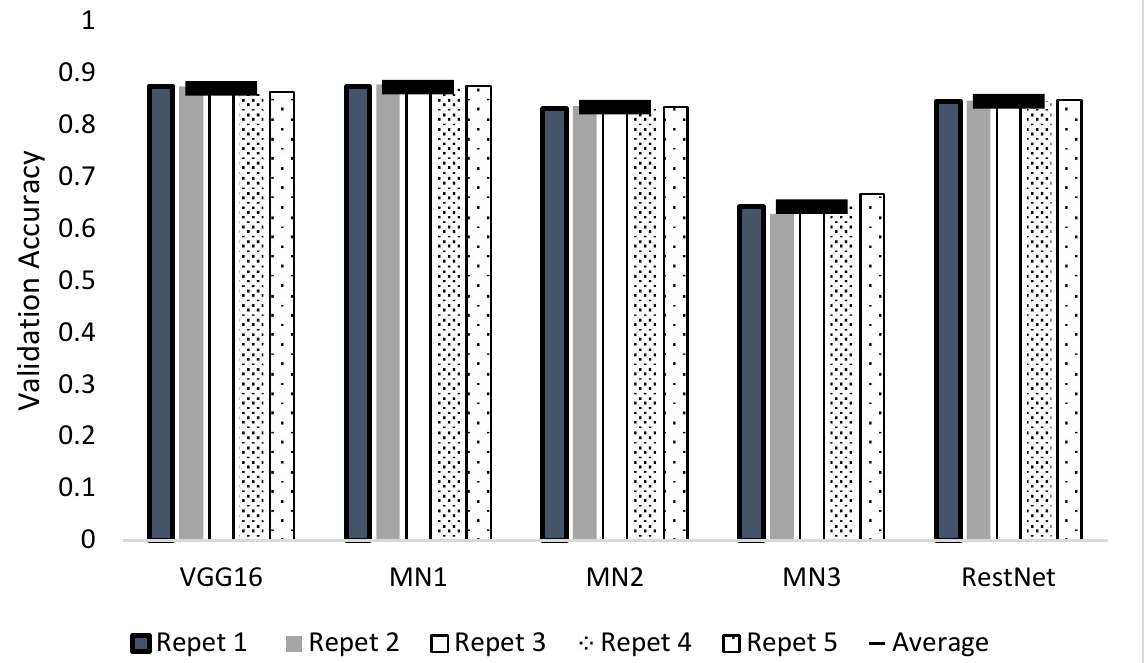}
\caption[center]{Sliding window fine tuning accuracy using 70\% of the dataset.Showing the four repeated experiment and the resulting average.}
\label{fig:D70_SW_rep}
\end{figure}

To validate our findings we have repeated the experiments for the block-wise and the SW approaches. The results shown in Figs \ref{fig:D70_Block_rep}, and \ref{fig:D70_SW_rep}, shows that both approaches yield consistent performance.

  \end{appendices}

%
%
%

%
%
%
%
%
%
%
%
%

\end{document}